\documentclass{bmvc2k}

\usepackage{paralist}
\usepackage{multirow}
\usepackage{amssymb}
\usepackage{pifont}
\usepackage[mathscr]{euscript}
\usepackage{cleveref}
\usepackage[skip=3pt plus1pt, indent=0pt]{parskip}
\crefformat{section}{\S#2#1#3} 
\crefformat{subsection}{\S#2#1#3}
\crefformat{subsubsection}{\S#2#1#3}
\usepackage{enumitem}
\setlist{topsep=0pt, leftmargin=*}

\title{Self-adversarial Multi-scale Contrastive Learning for Semantic Segmentation of Thermal Facial Images}

\addauthor{Jitesh Joshi}{jitesh.joshi.20@ucl.ac.uk}{1}
\addauthor{Nadia Bianchi-Berthouze}{nadia.berthouze@ucl.ac.uk}{1}
\addauthor{Youngjun Cho*}{youngjun.cho@ucl.ac.uk}{1*}

\addinstitution{
 Department of Computer Science\\
 University College London\\
 London, UK
}

\runninghead{Joshi, Berthouze, Cho}{SAM-CL Framework for Semantic Segmentation}

\def\repolink{\href{https://github.com/PhysiologicAILab/SAM-CL}{GitHub}} 

\def\labgithubpage{\href{https://github.com/PhysiologicAILab}{https://github.com/PhysiologicAILab}}
\begin{document}

\maketitle

\begin{abstract}
Segmentation of thermal facial images is a challenging task. This is because facial features often lack salience due to high-dynamic thermal range scenes and occlusion issues. Limited availability of datasets from unconstrained settings further limits the use of the state-of-the-art segmentation networks, loss functions and learning strategies which have been built and validated for RGB images. To address the challenge, we propose Self-Adversarial Multi-scale Contrastive Learning (SAM-CL) framework as a new training strategy for thermal image segmentation. SAM-CL framework consists of a SAM-CL loss function and a thermal image augmentation (TiAug) module as a domain-specific augmentation technique. We use the Thermal-Face-Database to demonstrate effectiveness of our approach. Experiments conducted on the existing segmentation networks (UNET, Attention-UNET, DeepLabV3 and HRNetv2) evidence the consistent performance gains from the SAM-CL framework. Furthermore, we present a qualitative analysis with UBComfort and DeepBreath datasets to discuss how our proposed methods perform in handling unconstrained situations.
\end{abstract}

\section{Introduction}
\label{sec:intro}
Thermal infrared imaging of human skin enables remote physiological sensing and affective, psychological states monitoring \cite{cho_physiological_2019, yang_graph-based_2022, cho_nose_2019}. Studies have indicated temperature patterns over specific facial regions as important psychophysiological signatures. For instance, temperature changes over the nostril region can be converted into breathing signals \cite{murthy_noncontact_2006, pereira_remote_2015, cho_robust_2017}. Another example is the nosetip temperature pattern associated with vasomotor activity which is related to mental stress states \cite{cho_nose_2019, cho_instant_2019}. Automated computational pipelines for processing thermal images therefore require identification of regions of interest (ROIs) which are either defined by a fixed bounding box or by an anatomical mask, with the latter being more appropriate for reliable extraction of physiological signals \cite{duarte_segmentation_2014}. The automated identification of an anatomical mask requires every pixel in a thermal image to be labelled according to its respective anatomical region.

\noindent Such semantic semantic segmentation task is particularly complex in thermal imaging. In comparison with typical RGB or gray-scale facial images, thermal facial images possess much less prominent facial features. This is because thermal images represent the temperature distribution over the skin surface, which is affected by dynamically varying physiological state as well as ambient temperature \cite{gade_thermal_2014, cho_robust_2017}. In addition, variations over the thermal surface are very low and occlusions such as forehead hairs and eye-glasses further make it challenging to reliably segment the ROIs in unconstrained settings.

\noindent The state-of-the-art in semantic segmentation has been under continuous progression following the foundational work of Fully Convolutional Networks (FCN) \cite{long_fully_2015}. The methodological contributions in the research of semantic segmentation can be broadly categorized into i) model architectures, ii)  loss functions or learning strategies, and iii) data augmentation techniques. Significant development has been made towards deep-learning architectures for semantic segmentation with some notable ones such as: UNET \cite{ronneberger_u-net_2015} and its variants \cite{siddique_u-net_2021}, the family of DeepLab networks \cite{chen_deeplab_2017, chen_deeplab_2018, chen_rethinking_2017}, HRNet \cite{sun_high-resolution_2019}, and the more recent transformer based approaches such as ``HRNet + OCR + SegFix'' \cite{yuan_segmentation_2021}. Pretrained network backbones such as ResNet \cite{he_deep_2016}, Xception \cite{chollet_xception_2017}, and HRNet \cite{sun_high-resolution_2019}, have further accelerated the progress owing to the availability of large-scale RGB datasets \cite{hoiem_pascal_2009, deng_imagenet_2009, lin_microsoft_2014}. On the other side, the widely used loss functions for semantic segmentation include softmax cross-entropy loss \cite{de_boer_tutorial_2005}, DICE loss \cite{sudre_generalised_2017} and region mutual information (RMI) loss \cite{zhao_region_2019}, among others. 

\noindent The existing challenges in semantic segmentation of thermal images include the lack of availability of large-scale bench-marking datasets. Furthermore, the studies validating the effectiveness of the segmentation networks, data-augmentation techniques and loss functions for RGB images have not sufficiently addressed the segmentation challenges of thermal images acquired in unconstrained settings. Specifically, data-augmentation techniques developed for RGB images \cite{shorten_survey_2019, buslaev_albumentations_2020} do not consider thermal ambient conditions and therefore not suitable to augment thermal images. In addition, thermal data is single channel, and the basic properties such as transparency in RGB change to opacity in thermal images (e.g. clear glass). As thermal infrared wavelength is not transmissive for most of the objects, occlusions are observed more frequently. In addition, the variations in thermal ambient conditions result in varying appearances and can not be related to the brightness variations in RGB images \cite{cho_deep_2018-1}.

\noindent This work addresses the challenge of training segmentation network with datasets of limited size using a novel self-adversarial multi-scale contrastive learning (SAM-CL) framework (\Cref{sec:proposed method}). SAM-CL framework introduces a SAM-CL loss function (\Cref{SAM-CL loss function}) and a thermal image augmentation (TiAug) module (\Cref{method: thermal augmentation module}), while utilizing existing segmentation networks. The TiAug module serves as domain specific augmentation, while the SAM-CL loss function provides enhanced supervision in learning inter-class separation and intra-class proximity in the presence of adversarial-attacks by TiAug. We compare the performance of SAM-CL framework with the existing segmentation loss functions, supervised contrastive learning (CL) \cite{wang_exploring_2021} and Generative Adversarial Network (GAN) based approaches for segmentation \cite{zhang_seggan_2018, xue_segan_2018} in \Cref{quantitative performance comparison}. Our contributions are:
\begin{itemize}[nosep,leftmargin=1em,labelwidth=*,align=left]
    \item Self-Adversarial Multi-scale Contrastive-Learning (SAM-CL) framework that introduces following to train existing segmentation networks:
    \begin{itemize}[nosep,leftmargin=1em,labelwidth=*,align=left]
        \item Self-adversarial multi-scale contrastive loss function for semantic segmentation, to efficiently achieve intra-class proximity and inter-class separation of the feature representations by utilizing the adversarial attacks generated by the TiAug module.
        \item a Thermal image augmentation (TiAug) module to generate representations of unconstrained thermal settings by applying domain-specific transformations to the thermal images acquired in controlled settings.
    \end{itemize}
    \item Performance benchmark with semantic segmentation of facial regions on Thermal Face Database \cite{kopaczka_fully_2018, kopaczka_thermal_2019} as well as qualitative analysis on UBComfort \cite{olugbade_toward_2021} and DeepBreath datasets \cite{cho_deepbreath_2017} demonstrating the performance gains by the proposed techniques.
\end{itemize}

\section{Related Work}
\textbf{Semantic Segmentation:} Segmentation networks largely follow encoder-decoder schemes \cite{noh_learning_2015, long_fully_2015, ronneberger_u-net_2015, milletari_v-net_2016, zhou_unet_2020}. To learn cross-pixel dependencies, several models apply attention mechanisms \cite{ba_multiple_2015} for semantic segmentation \cite{hu_squeeze-and-excitation_2018, oktay_attention_2018, fu_dual_2019, jin_ra-unet_2020, huang_ccnet_2019, zhuang_laddernet_2019}. Atrous convolutions along with pyramid pooling in \emph{DeepLab} networks \cite{chen_deeplab_2017, chen_deeplab_2018, chen_rethinking_2017} enable learning of the multi-scale features. \emph{HRNet}, as proposed in \cite{sun_high-resolution_2019}, shows performance gain on the semantic segmentation task, among other tasks, by making use of high-resolution and multilevel representations. Furthermore, the more recent development using transformer networks such as \emph{HRNet + OCR + SegFix} has achieved competitive performances on multiple benchmarking datasets \cite{yuan_segmentation_2021}. While increased network complexity and deeper layers prove effective for training models with large-scale RGB datasets, it is challenging to benefit from the same with a limited dataset size as typically observed in the case of thermal infrared imaging.

\noindent Existing semantic segmentation approaches for thermal images are not equipped to handle real-world scenarios, such as occlusion and varying thermal ambient conditions. One earlier study on occlusion removal in thermal imaging \cite{wong_eyeglasses_2013} proposes a modelling-based method using kernel principal component analysis for removing a specific occluding object (eye-glasses). This method requires the use of a registered color image to reconstruct the occluded thermal image, limiting its generalizability unless a large-scale dataset with pair of color-images and thermal-images is available. Large scale datasets allow capturing diverse representations, though acquiring a large scale dataset with thermal imaging and performing pixel wise annotations for semantic segmentation remains impractical. Furthermore, the thermal imaging datasets that are currently available with facial images have been acquired in highly controlled settings \cite{kopaczka_fully_2018, kopaczka_thermal_2019, abdrakhmanova_speakingfaces_2021, kowalski_high-resolution_2018}. It is therefore required to review the data augmentation techniques that can allow achieving robust performance in real-world scenarios.

\noindent\textbf{Image Augmentation Techniques:} The commonly used augmentation techniques include geometric transformations as well as learning or modelling based methods \cite{shorten_survey_2019}. While geometric transformations are relevant for thermal images, augmentation techniques pertaining to variations in brightness and contrast in RGB images cannot be directly mapped to thermal images. Among the learning based methods, GAN \cite{goodfellow_generative_2014} and self-adversarial training (SAT) \cite{goodfellow_explaining_2015} have shown promising performance. SimGAN as proposed in \cite{shrivastava_learning_2017} utilizes simulator generated images and a GAN to synthesize realistic augmented eye images. This method relies on the effectiveness of a simulator in synthesizing images, which may not generalize for different scenes and image modalities. YOLOv4 \cite{bochkovskiy_yolov4_2020} showed the effectiveness of SAT based augmentation technique called Fast Gradient Sign Method (FGSM) \cite{goodfellow_explaining_2015} in which an original image gets updated instead of the network weights in one forward pass, and this altered image is then used as an adversarial attack to improve the robustness of the trained model. A more recent work on localization of image forgery \cite{zhuo_self-adversarial_2022} also highlights the usefulness of FGSM based self-adversarial attacks in augmenting the data. While existing SAT approaches increase the robustness of the model for subtle changes in an image, they are insufficient in modelling range of real-world scenarios. Unlike gradient based update of images in SAT approaches, our proposed method (\Cref{method: thermal augmentation module}) models plausible variations of unconstrained settings in thermal images for adversarial attacks to enhance the robustness of the trained model in such settings.

\noindent\textbf{Loss Functions or Learning Strategies:} In addition to segmentation network and augmentation techniques, loss function or learning strategy plays a crucial role in achieving higher performance. For semantic segmentation tasks, cross-entropy loss \cite{de_boer_tutorial_2005} and weighted cross-entropy loss functions have been widely used \cite{jadon_survey_2020, asgari_taghanaki_deep_2021}. In cases of unequal class distribution, due to imbalanced distribution of pixels between semantic classes, focal loss \cite{lin_focal_2017} and DICE loss functions \cite{sudre_generalised_2017} have been reported to show better performance. In a more recent development, researchers proposed a loss function based on mutual information between pixels and semantic regions \cite{zhao_region_2019}, and showed substantial improvements on benchmarking datasets. Learning strategies such as GAN \cite{goodfellow_generative_2014, zhang_seggan_2018, cherian_sem-gan_2019, dong_automatic_2019, xue_segan_2018, zhang_ms-gan_2018} and CL \cite{wang_exploring_2021, zhang_looking_2021, zhao_contrastive_2021} have also shown to be effective for the segmentation task. A recent study using GAN \cite{muller_convolutional_2021} proposes multi-class segmentation approach, though, it is limited to handle only the occluding objects learnt at training time. The corner stone for the success of GAN as well as CL approaches is the availability of large scale datasets, which limits their effective deployment in thermal imaging. Our work takes inspiration from GAN as well as CL, however unlike the critic network in GAN, the auxiliary network in our approach does not compete with the generator (segmentation network). In addition, unlike the use of feature space for sampling anchors in CL, our approach uses predicted segmentation masks or logits as anchors.

\section{Proposed Method: Self-Adversarial Multi-scale Contrastive Learning (SAM-CL)}
\label{sec:proposed method}
\begin{figure}[htb]
    \centering
    \includegraphics[width=0.90\textwidth]{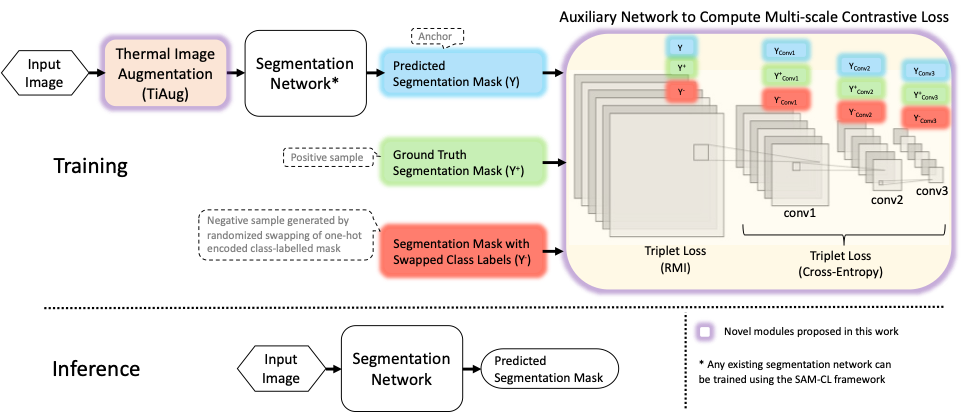}
    \caption{Proposed SAM-CL framework for Semantic Segmentation. In SAM-CL framework, the multi-scale contrastive-loss is computed by the four layer auxiliary network. TiAug generates augmented thermal images representing real-world scenarios.
    }
    \label{fig: SAM-CL framework architectural block diagram}
\end{figure}

Figure \ref{fig: SAM-CL framework architectural block diagram} provides an overview of our proposed Self-Adversarial Multi-scale Contrastive Learning (SAM-CL) framework. One of the highlights of our framework is that it is used only during the training, resulting in no computation overhead during the inference. 
Though this framework can be generically applied to segmentation tasks, the key objective in this work is to train a segmentation network on thermal facial images for: i) segmentation of facial regions including eyes, eyebrows, nose, mouth and chin area and, ii) resilience to varying thermal ambient conditions as well as occlusions in unconstrained settings. The proposed SAM-CL framework utilizes existing segmentation networks and introduces a SAM-CL loss function and a thermal image augmentation (TiAug) module to achieve the objective without requiring thermal images acquired in unconstrained scenarios.

\subsection{Loss Function}
\label{Loss function}
\subsubsection{Preliminaries}
\label{Loss function preliminaries}
For a semantic segmentation task, the segmentation network \(SEG\) learns a function \(f_{SEG}(I)\) that maps input image \(I\) to the ground-truth mask \(Y\), that specifies a semantic class \(c \in C\) for every pixel \(i \in I\). 
A limitation of the most commonly used pixel-wise segmentation loss functions such as cross-entropy loss, is their inability to capture the relationships between pixels. To address this limitation, a recent work proposed the mutual information based loss function \cite{zhao_region_2019} that combines cross-entropy and structural similarity loss.
Furthermore, to learn the relationship between pixels of multiple images and to supervise the representations within pixel-embedding, supervised contrastive loss for semantic segmentation is proposed in \cite{wang_exploring_2021}:
\begin{equation}\label{embeddings based contrastive loss}
\mathscr{L}_{y}^{NCE} 
= 
\frac{1}{|P_{y}|} \sum_{y^{+} \in P_{y}}
-log \frac{\exp({y \cdot y^{+}/\tau)}}{\exp({y \cdot y^{+}/\tau}) + \sum_{y^{-} \in N_{y} } \exp({y \cdot y^{-}/\tau})}
\end{equation}
where \(P_{y}\) and \(N_{y}\) are positive and negative samples of pixel-embedding, belonging to classes dissimilar to that of the anchor pixel \(y\). This learning strategy is very effective in maximizing inter-class separation, while minimizing intra-class distance within pixel-embedding, specially when large-scale dataset and corresponding pretrained weights are available. However, feature representations within pixel-embeddings remain transient while training a segmentation network without pretrained weights and with a limited dataset size. This limits the effectiveness of CL in training the segmentation network. To address this, we resort to a CL strategy that uses logits instead of pixel-embedding, while remaining effective in maximizing inter-class separation and minimizing intra-class proximity. 

\subsubsection{SAM-CL Loss Function}
\label{SAM-CL loss function}
In a one-hot encoded ground-truth segmentation mask (\(Y^{+}_{oh}\)), each channel represents a binary mask for the respective classes. Class swapped mask (\(Y^{-}_{oh}\)) is generated by randomized swapping of channels of the (\(Y^{+}_{oh}\)) with the constraint that no channels of \(Y^{+}_{oh}\) and \(Y^{-}_{oh}\) match. With logits or one-hot predicted mask \(Y_{oh}\) representing an anchor, \(Y^{+}_{oh}\) and \(Y^{-}_{oh}\) representing positive and negative samples respectively, the first triplet loss is computed as shown in \Cref{eq: Triplet loss}:
\begin{equation}\label{eq: Triplet loss}
\mathscr{L}_{s0}(Y_{oh}, Y^{+}_{oh}, Y^{-}_{oh}) =
\max \{d(Y_{oh}, Y^{+}_{oh}) - d(Y_{oh}, Y^{-}_{oh}) + {\textrm margin}, 0\}
\end{equation}
\Cref{eq: Triplet loss} allows learning inter-class separation as well as intra-class proximity without requiring to compute the contrastive loss with pixel-embedding. As \(Y^{-}_{oh}\) preserves spatial features at mask-level, the optimization results in effective inter-class separation of the spatial features. \(Y_{oh}\), \(Y^{+}_{oh}\), and \( Y^{-}_{oh}\) are passed through a 4-layered auxiliary network in three different forward passes to compute the feature maps \(y_{Conv1}\), \(y_{Conv2}\) and \(y_{Conv3}\); (\(y = Y_{oh}, Y^{+}_{oh}, Y^{-}_{oh}\)) for each layer. Down-scaling of 2 is applied at each layer with the number of channels in every layer held constant and equal to the number of classes, consistent with the first layer input channels.
\begin{multline}\label{eq: SAM-CL loss function}
\mathscr{L}_{SAM-CL} = \mathscr{L}_{s0}(Y_{oh}, Y^{+}_{oh}, Y^{-}_{oh}) +
\mathscr{L}_{s1}(Y_{Conv1}, Y^{+}_{Conv1}, Y^{-}_{Conv1}) + \\
\mathscr{L}_{s2}(Y_{Conv2}, Y^{+}_{Conv2}, Y^{-}_{Conv2}) +
\mathscr{L}_{s3}(Y_{Conv3}, Y^{+}_{Conv3}, Y^{-}_{Conv3})
\end{multline}
The overall SAM-CL loss function as formulated in \Cref{eq: SAM-CL loss function}, therefore offers supervision to maximize inter-class separation at multiple-scales. To compute the distance function \(d(x,y)\) as mentioned in \Cref{eq: Triplet loss}, we use RMI \cite{zhao_region_2019} on logits, and cross-entropy loss for down-convolved feature-maps in the auxiliary network (see \Cref{fig: SAM-CL framework architectural block diagram}). 

\subsection{Thermal Image Augmentation (TiAug) Module}
\label{method: thermal augmentation module}
\begin{figure}[htb]
    \centering
    \includegraphics[width=0.9\textwidth]{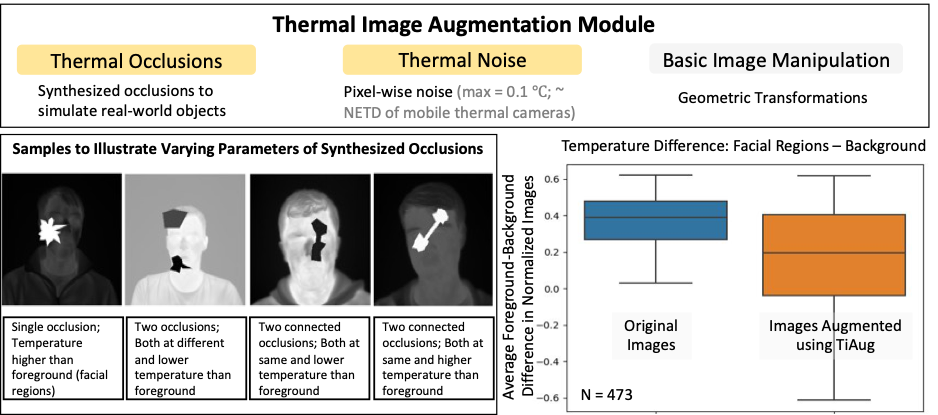}
    \caption{Thermal Image Augmentation (TiAug) Module. TiAug module consists of modules for thermal data specific occlusion generation as well as additive thermal noise modules along with the commonly used geometric transformation modules.}
    \label{fig: thermal augmentation module}
\end{figure}

The thermal image augmentation module (TiAug), as illustrated in \Cref{fig: thermal augmentation module}, transforms a thermal image acquired in controlled settings into an image resembling one acquired in unconstrained ambient settings. This is an important step as there often exist high-dynamic thermal range scenes in the real world settings \cite{cho_robust_2017}. Inspired by the Optimal Quantisation technique in  \cite{cho_robust_2017}, this module is designed to first add synthesized objects with diverse parameters in an occluding as well as a non-occluding manner. These parameters include size, shape, temperature, position and configuration (i.e. single, dual or dual-connected objects). In addition, a random temperature value is added as thermal noise to every pixel. The maximum magnitude of the noise is set as per the noise equivalent temperature difference (NETD), a sensitivity parameter of thermal infrared imaging camera, that provides the minimum value of temperature difference that can be sensed reliably by a camera. While a high-sensitive thermal camera has lower magnitude of NETD, it is higher for the low-cost mobile thermal imaging camera. TiAug sets the maximum NETD value (\(Th^{max}_{NETD} = 0.1^{\circ}C\)) considering the low-cost thermal imaging camera.

\noindent As expressed in \Cref{eq: TiAug transformation}, an augmented image (\(I_{aug}^{HxW}\)) is generated from an original image (\(I_{org}^{HxW}\)) of height (\(H\)) and width (\(W\)) by applying an occlusion transformation (\(f_{occ}\)) and adding thermal noise (\(\eta^{HxW}\): \(0 < \kappa < Th^{max}_{NETD} \forall \kappa \in \eta^{HxW}\)). The parameters characterizing occluding objects in \(f_{occ}\) are size (\(\vartheta_{sz}\): 1-40\% of facial region), shape (\(\vartheta_{sh}\)) characterized by number of vertices (2-30) and the corresponding irregularity (0 to 10\%) as well as spikeness (0 to 50\%), temperature (\(\vartheta_{temp}\): \(I_{org}^{min}-(I_{org}^{max}-I_{org}^{min}) \leq \vartheta_{temp} \leq I_{org}^{max}+(I_{org}^{max}-I_{org}^{min}\))), position (\(\vartheta_{xy}\): \((x,y) \in R^{HxW}\)), and configurations (\(\vartheta_{config}\): single, dual and dual-connected). The function \(g(...)\) uniquely combines the values of each parameter from the respective range, allowing the synthesis of endless new representations. \(f_{occ}\) replaces the specific pixels of \(I_{org}^{HxW}\) with the synthesized objects which may or may not occlude the facial regions. This leads to real-world histogram distribution of thermal images, unlike bimodal distribution when thermal images are acquired in controlled settings. In a bimodal histogram distribution, a peak at a lower temperature value depicts background and a peak at a higher temperature corresponds to the facial regions. Several segmentation algorithms assume bimodal histogram distribution for an automated segmentation \cite{isalkar_analysis_2022}, which affects their performance in unconstrained settings. 
\begin{equation}\label{eq: TiAug transformation}
I_{aug}^{HxW} = f_{occ}(I_{org}^{HxW}, g(\vartheta_{sz}, \vartheta_{sh}, \vartheta_{temp}, \vartheta_{xy}, \vartheta_{config})) + \eta^{HxW}
\end{equation}

\noindent Real-world scenarios may include objects at temperatures higher than the facial regions (e.g., sun, hot beverages), as well as objects at lower temperatures \cite{cho_robust_2017}. In such real-world scenarios, histogram distribution is not generally bimodal. To simulate real-world variations in histogram distribution pattern, TiAug adds synthesized objects both at temperatures higher and lower than that of the facial regions. This approach prevents the deep-learning network from over-fitting to bimodal-distributions. Hence, the obtained \({I}_{aug}^{HxW}\) represents real-world scenarios both in terms of spatial characteristics of ambient objects as well as the histogram distribution of temperature values. \({I}_{aug}^{HxW}\) is further normalized and passed as input to the segmentation network. While the examples in the \Cref{fig: thermal augmentation module} show changes in spatial characteristics, the box-plot analysis signifies the effectiveness of TiAug in altering average foreground (facial-regions) temperature and average background temperature in the normalized images. Furthermore, TiAug applies the following geometric transformations: horizontal flip, vertical flip, rotation, Gaussian blur, and resizing (0.5X to 2X).

\section{Experiments}
\label{experiment}
We perform various experiments to compare the proposed SAM-CL framework with existing loss functions and learning strategies, using the following segmentation networks: U-NET \cite{ronneberger_u-net_2015}, Attention UNET \cite{oktay_attention_2018}, DeepLabV3 \cite{chen_encoder-decoder_2018} and HRNet \cite{sun_high-resolution_2019}. Our code uses PyTorch \cite{paszke_pytorch_2019} and is available on \repolink{}. It is built upon a prior work on contrastive learning for semantic segmentation \cite{wang_exploring_2021}. Given a lack of benchmark segmentation performance reports on thermal facial datasets as well as pretrained models, we implement and train the aforementioned prior-art segmentation networks with Xavier uniform initialization. We use a batch size of 16 along with an SGD optimiser with a weight decay of 1e-8 and betas set to 0.9 and 0.999.

\noindent For our experiments, it is required to use datasets of raw thermal matrices (ie. absolute temperature value assigned to each pixel). The available datasets of thermal facial images include: Thermal Face Database \cite{kopaczka_fully_2018, kopaczka_modular_2019}, UBComfort dataset \cite{olugbade_toward_2021}, and DeepBreath dataset \cite{cho_deepbreath_2017, cho_robust_2017}. For training and quantitative evaluation (\Cref{qualitative analysis}), we mainly use the Thermal Face Database \cite{kopaczka_fully_2018, kopaczka_modular_2019} as it provides, together with the data, the ground-truth labels. A limitation of this dataset is that it is acquired in controlled setting. For this reason, we have use the other two datasets collected in uncontrained settings. However, due to the lack of ground truth in these two datasets, we use them for qualitative analysis only (\Cref{qualitative analysis}) to demonstrate the effectiveness of SAM-CL framework in such settings.

\subsection{Quantitative Evaluation on Thermal Face Database}
\label{quantitative evaluation}
\subsubsection{Dataset Description}
\label{thermal dataset description}
Thermal Face Database \cite{kopaczka_fully_2018, kopaczka_modular_2019} consists of 2935 images of 90 individuals with 68 manually annotated facial landmark points. We derive segmentation masks from the landmarks points for each anatomical region (chin, mouth, nose, eyes and eye-brows). Data is split into training (85\%) and validation (15\%) sets based on subject ids.

\subsubsection{Results}
\label{quantitative performance comparison}
To investigate the generalizability of the SAM-CL framework, we train UNET \cite{ronneberger_u-net_2015}, Attention UNET \cite{oktay_attention_2018}, DeepLabV3 \cite{chen_encoder-decoder_2018} and HRNet \cite{sun_high-resolution_2019} segmentation networks. In all the experiments, augmentation with basic geometric transformations including horizontal flip, vertical flip, rotation, Gaussian blur, and resizing (0.5X to 2X), is uniformly applied. The loss functions used for bench-marking includes weighted binary cross-entropy loss (BCE), DICE loss, and region mutual information (RMI) loss \cite{zhao_region_2019}. As SAM-CL framework relates to CL and GAN, we additionally compare performance with SegAN \cite{xue_segan_2018, xue_adversarial_2018}, SegGAN \cite{zhang_seggan_2018}, along with a recent work on the supervised CL applied to semantic segmentation \cite{wang_exploring_2021}.
\begin{table}[htb]
\centering
\caption{Performance Evaluation of SAM-CL Framework\\}
\label{tab: performance evaluation of SAM-CL framework}
\renewcommand{\arraystretch}{1.60}
\resizebox{0.90\textwidth}{!}{%
\begin{tabular}{llll|llll}
\cline{1-3} \cline{6-8}
\multicolumn{1}{c}{\textbf{Segmentation Network}} & \textbf{\begin{tabular}[c]{@{}l@{}}Learning Strategy\\ (Loss Function)\end{tabular}} & \textbf{mIoU (\%)} &  &  & \multicolumn{1}{c}{\textbf{Segmentation Network}} & \textbf{\begin{tabular}[c]{@{}l@{}}Learning Strategy\\ (Loss Function)\end{tabular}} & \textbf{mIoU (\%)} \\ \cline{1-3} \cline{6-8} 
\multirow{7}{*}{UNET \cite{ronneberger_u-net_2015}} & Pixel-wise Segmentation (BCE) & 67.64 &  &  & \multirow{7}{*}{Attention UNET \cite{oktay_attention_2018}} & Pixel-wise Segmentation (BCE) & 66.61 \\ \cline{2-3} \cline{7-8} 
 & Pixel-wise Segmentation (DICE) & 75.00 &  &  &  & Pixel-wise Segmentation (DICE) & 75.14 \\ \cline{2-3} \cline{7-8} 
 & GAN (SegAN) \cite{xue_segan_2018} & 76.79 &  &  &  & GAN (SegAN) \cite{xue_segan_2018} & 76.75 \\ \cline{2-3} \cline{7-8} 
 & GAN (SegGAN) \cite{zhang_seggan_2018} & 75.50 &  &  &  & GAN (SegGAN) \cite{zhang_seggan_2018} & 76.24 \\ \cline{2-3} \cline{7-8} 
 & RMI \cite{zhao_region_2019} & 81.35 &  &  &  & RMI \cite{zhao_region_2019} & 81.39 \\ \cline{2-3} \cline{7-8} 
 & ContrastiveSeg \cite{wang_exploring_2021} & 81.24 &  &  &  & ContrastiveSeg \cite{wang_exploring_2021} & 81.50 \\ \cline{2-3} \cline{7-8} 
 & \textbf{SAM-CL (Ours)} & \textbf{82.11 (+0.76)} &  &  &  & \textbf{SAM-CL (Ours)} & \textbf{82.85 (+1.35)} \\ \cline{1-3} \cline{6-8} 
\multirow{3}{*}{DeepLabV3+ResNet101 \cite{chen_deeplab_2018, he_deep_2016}} & RMI \cite{zhao_region_2019} & 75.85 &  &  & \multirow{3}{*}{HRNetV2-W48 \cite{sun_high-resolution_2019}} & RMI \cite{zhao_region_2019} & 78.46 \\ \cline{2-3} \cline{7-8} 
 & ContrastiveSeg \cite{wang_exploring_2021} & 74.45 &  &  &  & ContrastiveSeg \cite{wang_exploring_2021} & 78.36 \\ \cline{2-3} \cline{7-8} 
 & \textbf{SAM-CL (Ours)} & \textbf{79.29 (+3.44)} &  &  &  & \textbf{SAM-CL (Ours)} & \textbf{78.97 (+0.61)} \\ \cline{1-3} \cline{6-8} 
\end{tabular}%
}
\end{table}

\noindent Table \ref{tab: performance evaluation of SAM-CL framework} shows the comparison of performances with percentage mean IoU metric. We observe consistent performance gains with the use of DICE loss when compared against BCE loss. GAN based learning strategy \cite{xue_segan_2018, zhang_seggan_2018} is found effective for UNET and Attention UNET, while the performance drops for DeepLabV3 network when comparing against the respective performance with the DICE loss. Consistent performance improvements from DICE loss function as well as GAN based learning strategy is evident for the models trained with RMI loss function. The performance gains across all the segmentation networks can be noted when deploying the SAM-CL framework.
\begin{table}[htb]
\centering
\caption{Ablation Study for TiAug and SAM-CL Loss Function}
\label{tab: ablation study}
\renewcommand{\arraystretch}{1.35}
\resizebox{0.6\textwidth}{!}{%
\begin{tabular}{lccc}
\hline
\multirow{2}{*}{\textbf{Segmentation Network}} & \multicolumn{3}{c}{\textbf{mIoU (\%) Performance}} \\ \cline{2-4} 
 & \multicolumn{1}{l}{\textbf{RMI}} & \multicolumn{1}{l}{\textbf{RMI + TiAug}} & \multicolumn{1}{l}{\textbf{RMI + TiAug + SAM-CL}} \\ \hline
UNET \cite{ronneberger_u-net_2015} & 81.36 & 81.91 & \textbf{82.11} \\ \hline
Attention UNET \cite{oktay_attention_2018} & 81.39 & 82.29 & \textbf{82.85} \\ \hline
HRNetV2-W48 \cite{sun_high-resolution_2019} & 78.13 & 78.87 & \textbf{78.97} \\ \hline
DeepLabV3+ResNet101 \cite{chen_deeplab_2018, he_deep_2016} & 75.85 & 78.07 & \textbf{78.12} \\ \hline
DeepLabV3+Xception \cite{chen_deeplab_2018, chollet_xception_2017} & 76.55 & 77.31 & \textbf{77.85} \\ \hline
\end{tabular}%
}
\end{table}

\noindent We performed an ablation study to examine the individual contribution of the TiAug module and the SAM-CL loss function. From \Cref{tab: ablation study}, we observe performance gains from the baseline, trained with RMI loss, for all the segmentation networks when the data is augmented using the TiAug module. Similarly additional performance gains for the respective segmentation networks are observed when the SAM-CL loss function is used along with the TiAug module. The TiAug module presents a segmentation network with the adversaries such as occlusions and varying ambient temperature levels in the input thermal images, while the SAM-CL loss function maximizes the inter-class separation using the class-swapped negative sample \(Y^{-}\) and its down-scaled representations in the auxiliary network.

\subsection{Qualitative Analysis}
\label{qualitative analysis}
\begin{figure}[htb]
    \centering
    \includegraphics[width=0.95\textwidth]{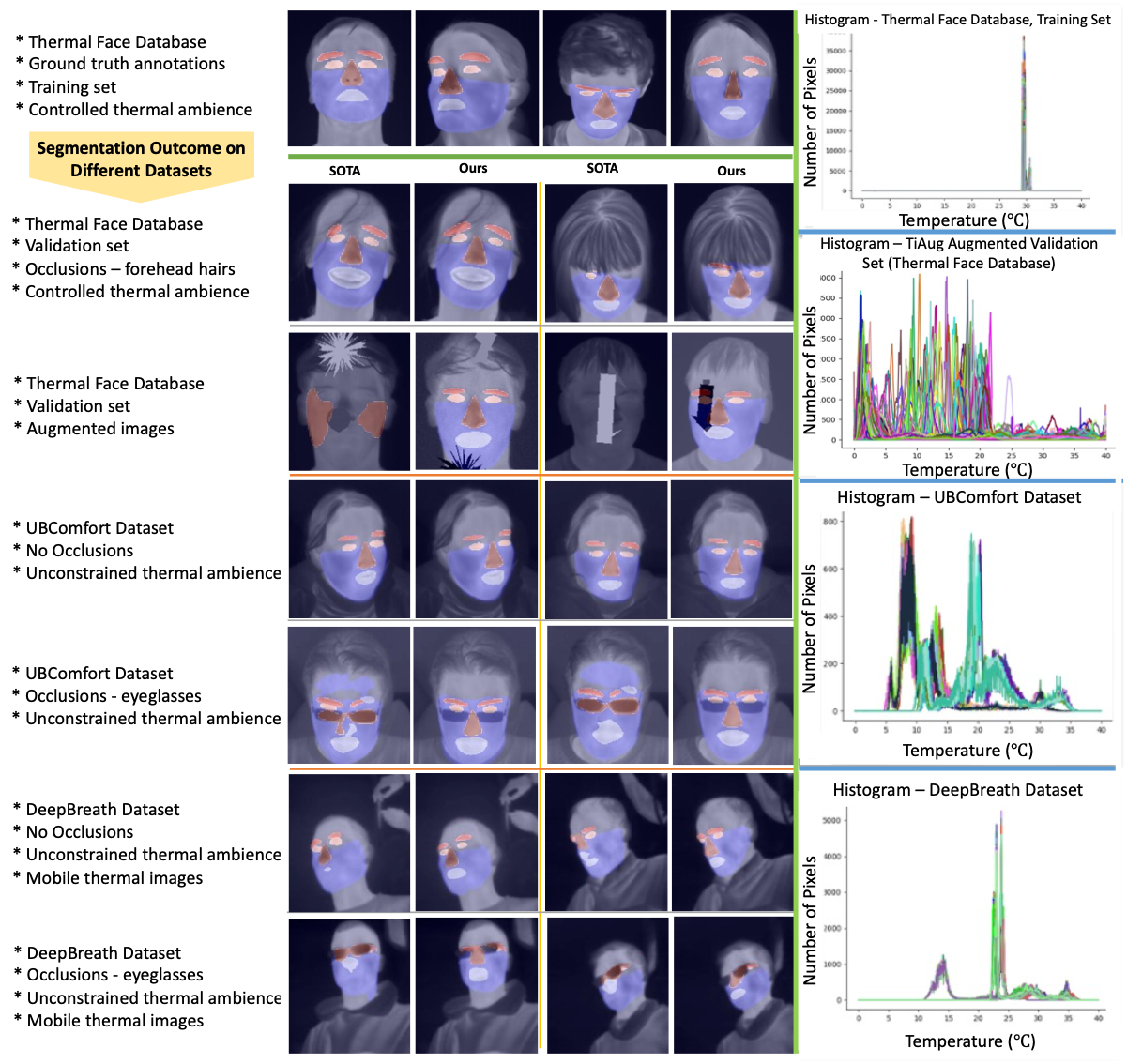}
    \caption{Qualitative Performance Analysis on Different Thermal Imaging Datasets. Attention UNET is trained using RMI loss to compare the performance of the proposed SAM-CL framework. Please refer to \Cref{qualitative analysis} for the discussion on qualitative performance analysis.} 
    \label{fig:qualitative result synthesized}
\end{figure}

As Thermal Face Database does not include real-world occlusions, we extend the evaluation of our approach with a qualitative analysis on the datasets acquired in unconstrained settings (see \Cref{fig:qualitative result synthesized}). The UBComfort dataset \cite{olugbade_toward_2021} was acquired from in-the-wild car users with varying thermal ambient conditions using a high-resolution thermal camera. 
Instead, the DeepBreath dataset \cite{cho_deepbreath_2017} was acquired from participants under induced stress levels using a low-resolution mobile thermal camera.

\noindent The thermal images in the first row of the \Cref{fig:qualitative result synthesized} present ground-truth labelmask overlaid with color-coded class-labels. The histogram plot on the top-right shows the temperature distribution across the images in the Thermal Face database, highlighting the highly controlled laboratory settings. We further identify a few samples (row-2) within the Thermal Face Database in which hairs occlude a small part of the thermal image. The superiority of the model trained jointly with the SAM-CL loss function and the TiAug module in reliably handling forehead hairs occlusions is evidenced by the segmentation outcome. In the following row, we present samples generated by the TiAug module, which after min-max normalization, appear significantly different, as would be the case when objects that are either too hot or too cold appear in an image. The state-of-the-art (SOTA) model fails in these scenarios (row-3) as it has not been trained with such variations. 

\noindent The thermal images of individuals without and with eye-glasses, seated in a car \cite{olugbade_toward_2021} are shown in rows 4 and 5 of \Cref{fig:qualitative result synthesized} respectively. Though the training set does not include such images, the model trained with our proposed method shows resilience towards performing reliable segmentation in the presence of eye-glasses, while the SOTA sub-performs. Similarly, the thermal images of individuals performing cognitive tasks \cite{cho_deepbreath_2017}, without and with eye-glasses, are shown in rows 6 and 7 respectively. While SOTA sub-performs on both cases, SAM-CL framework shows reliable performance. It can be noted that the thermal images in the DeepBreath dataset \cite{cho_deepbreath_2017} are acquired using mobile thermal camera (FLIR One), highlighting the robustness of the model trained using SAM-CL framework across for different thermal camera specifications.

\section{Conclusion}
The proposed SAM-CL framework that introduces the SAM-CL loss function and the TiAug module, is shown to be effective in training segmentation networks with datasets of limited size. The TiAug module transforms thermal images acquired in controlled laboratory environment into ones representing real-world scenarios. This transformation considers a range of plausible ambient temperature, geometric properties of common occluding objects as well as noise specification of widely used thermal cameras. This makes the TiAug module suitable to be applied on thermal images for various learning-based computer-vision tasks including classification, object detection, instance and panoptic segmentation to train deep-learning networks to handle common real-world scenarios, without explicitly requiring a thermal dataset to be acquired in such scenarios. 

\noindent Furthermore, the SAM-CL loss function benefits from the TiAug module that presents the segmentation network with adversaries (e.g. occluded images), resulting in a portion of predicted logits to overlap with the synthesized negative sample (from class-swapped labelmask). This overlap of logits with negative sample results in a higher loss for the incorrect label predicted by the segmentation network in the corresponding region. This explains the effectiveness of the SAM-CL loss function in conjunction with the TiAug module in offering consistent performance gains across different segmentation networks. SAM-CL loss function can be extended to other imaging modalities to train segmentation networks with datasets of limited size, by devising appropriate augmentation technique for self-adversarial learning.

\noindent\textbf{Acknowledgements:} Jitesh Joshi was fully supported by Physiological Computing Studentship from UCL Computer Science (\labgithubpage). 

\bibliography{main}
\end{document}